\documentclass[11pt,a4paper]{article}
\usepackage[hyperref]{naaclhlt2018}
\usepackage{times}
\usepackage{latexsym}

\usepackage[T2A, T1]{fontenc}
\usepackage[utf8]{inputenc}

\usepackage{url}
\usepackage{graphicx}
\usepackage{multirow}
\usepackage{amsthm}
\usepackage{amssymb}
\usepackage{amsmath}
\usepackage{amsfonts}
\usepackage{color}
\usepackage{mdwlist}
\usepackage{microtype}
\usepackage{tabularx}
\usepackage{booktabs}
\usepackage{subcaption}
\usepackage{mwe} 
\DeclareCaptionLabelFormat{andtable}{#1˜#2 \& \tablename˜\thetable}

\usepackage{color,colortbl}

\definecolor{Gray}{gray}{0.9}
\usepackage{enumitem,amssymb}
\newlist{todolist}{itemize}{2}
\setlist[todolist]{label=$\square$}
\usepackage{pifont}

\newcolumntype{b}{X}
\newcolumntype{m}{>{\hsize=.6\hsize}X}
\newcolumntype{s}{>{\hsize=.33\hsize}X}

\makeatletter
\newcommand{\removelatexerror}{\let\@latex@error\@gobble}

\makeatother

\usepackage[russian,english]{babel}

\aclfinalcopy 

\title{Specialising Word Vectors for Lexical Entailment}

\author{{Ivan Vuli\'c}$^{\mathbf{1}}$ \and Nikola Mrk\v{s}i\'{c}$^{\mathbf{2}}$\\
$^{\mathbf{1}}$ Language Technology Lab, University of Cambridge \\
$^{\mathbf{2}}$ PolyAI\\
\texttt{iv250@cam.ac.uk} \hspace{1.5em} \texttt{nikola@poly-ai.com}\\
}

\date{}

\begin{document}
\maketitle
\begin{abstract}
We present \textsc{lear} (\textbf{L}exical \textbf{E}ntailment \textbf{A}ttract-\textbf{R}epel), a novel post-processing method that transforms any input word vector space to emphasise the asymmetric relation of \textit{lexical entailment} (\textsc{LE}), also known as the \textsc{is-a} or hyponymy-hypernymy relation. By injecting external linguistic constraints (e.g., WordNet links) into the initial vector space, the \textsc{LE} specialisation procedure brings true hyponymy-hypernymy pairs closer together in the transformed Euclidean space. The proposed asymmetric distance measure adjusts the norms of word vectors to reflect the actual WordNet-style hierarchy of concepts. Simultaneously, a joint objective enforces semantic similarity using the symmetric cosine distance, yielding a vector space specialised for both lexical relations at once. \textsc{lear} specialisation achieves state-of-the-art performance in the tasks of hypernymy directionality, hypernymy detection, and graded lexical entailment, demonstrating the effectiveness and robustness of the proposed asymmetric specialisation model.
\end{abstract}

\section{Introduction}
\label{s:intro}
Word representation learning has become a research area of central importance in NLP, with its usefulness demonstrated across application areas such as parsing \cite{Chen:2014emnlp}, machine translation \cite{Zou:2013emnlp}, and many others \cite{Turian:2010acl,Collobert:2011jmlr}. Standard techniques for inducing word embeddings rely on the \textit{distributional hypothesis} \cite{Harris:1954}, using co-occurrence information from large textual corpora to learn meaningful word representations \cite
{Mikolov:2013nips,Levy:2014acl,Pennington:2014emnlp,Bojanowski:2017tacl}. %

A major drawback of the distributional hypothesis is that it coalesces different relationships between words, such as synonymy and topical relatedness, into a single vector space. A popular solution is to go beyond stand-alone unsupervised learning and fine-tune distributional vector spaces by using external knowledge from human- or automatically-constructed knowledge bases. This is often done as a \textit{post-processing} step, where distributional vectors are gradually refined to satisfy linguistic constraints extracted from lexical resources such as WordNet \cite{Faruqui:2015naacl,Mrksic:2016naacl}, the Paraphrase Database (PPDB) \cite{Wieting:2015tacl}, or BabelNet \cite{Mrksic:2017tacl,Vulic:2017emnlp}. One advantage of post-processing methods is that they treat the input vector space as a \textit{black box}, making them applicable to any input space.

A key property of these methods is their ability to transform the vector space by \textit{specialising} it for a particular relationship between words.\footnote{Distinguishing between synonymy and antonymy has a positive impact on real-world language understanding tasks such as Dialogue State Tracking \cite{Mrksic:2017acl}.} Prior work has predominantly focused on distinguishing between semantic similarity and conceptual relatedness \cite{Faruqui:2015naacl,Mrksic:2017tacl,Vulic:2017acl}. In this paper, we introduce a novel post-processing model which specialises vector spaces for the \textit{lexical entailment} (\textsc{le}) relation.


Word-level lexical entailment is an \textit{asymmetric} semantic relation \cite{Collins:1972article,Beckwith:1991wn}. It is a key principle determining the organisation of semantic networks into hierarchical structures such as semantic ontologies \cite{Fellbaum:1998wn}. Automatic reasoning about \textsc{le} supports tasks such as taxonomy creation \cite{Snow:2006acl,Navigli:2011ijcnlp}, natural language inference \cite{Dagan:2013book,Bowman:2015emnlp}, text generation \cite{Biran:2013ijcnlp}, and metaphor detection \cite{Mohler:2013ws}. 

\begin{figure}
\centering
\includegraphics[width=0.9\linewidth]{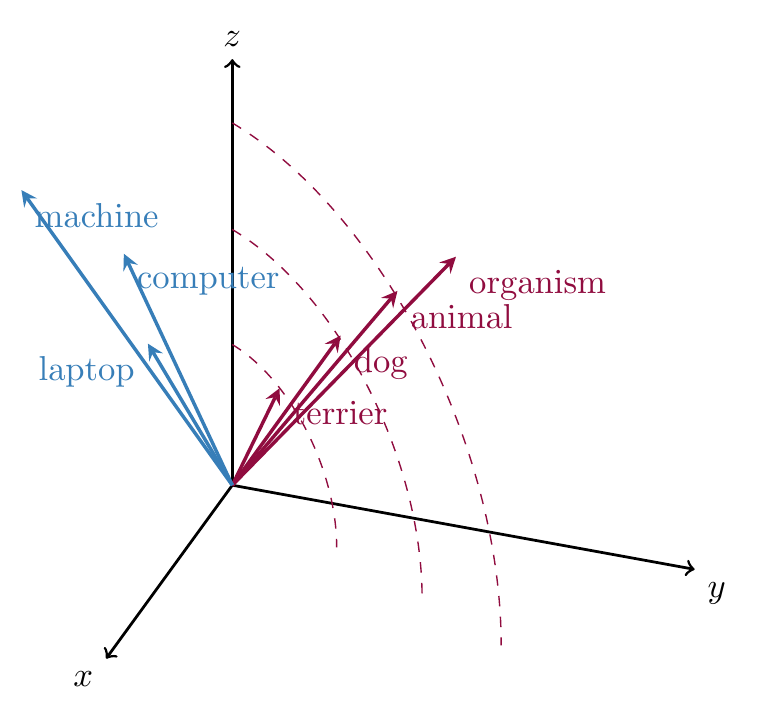}
\vspace{-0.5em}
\caption{An illustration of \textsc{lear} specialisation. \textsc{lear} controls the arrangement of vectors in the transformed vector space by: \textbf{1)} emphasising symmetric similarity of \textsc{le} pairs through cosine distance (by enforcing small angles between $\overrightarrow{terrier}$ and $\overrightarrow{dog}$ or $\overrightarrow{dog}$ and $\overrightarrow{animal}$); and \textbf{2)} by imposing an \textsc{le} ordering using vector norms, adjusting them so that higher-level concepts have larger norms (e.g., $|\overrightarrow{animal}| > |\overrightarrow{dog}| > |\overrightarrow{terrier}|$).}
\label{fig:vectors}
\vspace{-0.0em}
\end{figure}

Our novel \textsc{le} specialisation model, termed \textsc{lear} (\textbf{L}exical \textbf{E}ntailment \textbf{A}ttract-\textbf{R}epel), is inspired by \textsc{Attract-Repel}, a state-of-the-art general specialisation framework \cite{Mrksic:2017tacl}.\footnote{https://github.com/nmrksic/attract-repel} The key idea of \textsc{lear}, illustrated by Figure~\ref{fig:vectors}, is to pull
desirable (\textsc{attract}) examples described by the constraints closer together, while at the same time pushing undesirable (\textsc{repel}) word pairs away from each other. Concurrently, \textsc{lear} (re-)arranges vector norms so that norm values in the Euclidean space reflect the hierarchical organisation of concepts according to the given \textsc{le} constraints: put simply, higher-level concepts are assigned larger norms. Therefore, \textsc{lear} simultaneously captures the hierarchy of concepts (through vector norms) and their similarity (through their cosine distance). The two pivotal pieces of information are combined into an \textit{asymmetric distance measure} which quantifies the \textsc{LE} strength in the specialised space.

After specialising four well-known input vector spaces with \textsc{lear}, we test them in three standard word-level \textsc{le} tasks \cite{Kiela:2015acl}: \textbf{1)} hypernymy \textit{directionality}; \textbf{2)} hypernymy \textit{detection}; and \textbf{3)} \textit{combined} hypernymy detection/directionality. Our specialised vectors yield notable improvements over the strongest baselines for each task, with each input space, demonstrating the effectiveness and robustness of \textsc{lear} specialisation.

The employed asymmetric distance allows one to make graded assertions about hierarchical relationships between concepts in the specialised space. This property is evaluated using HyperLex, a recent \textit{graded LE} dataset \cite{Vulic:2017cl}. The \textsc{lear}-specialised vectors  push state-of-the-art Spearman's correlation from 0.540 to 0.686 on the full dataset (2,616 word pairs), and from 0.512 to 0.705 on its noun subset (2,163 word pairs).

The code for the \textsc{lear} model is available from: \url{github.com/nmrksic/lear}.

\section{Methodology} 
\label{s:methodology}
\subsection{The \textsc{Attract-Repel} Framework}

Let $V$ be the vocabulary, $A$ the set of \textsc{Attract} word pairs (e.g.,~\emph{intelligent} and \emph{brilliant}), and $R$ the set of \textsc{Repel} word pairs (e.g.,~\emph{vacant} and \emph{occupied}). The \textsc{Attract-Repel} procedure operates over mini-batches of such pairs $\mathcal{B}_{A}$ and $\mathcal{B}_{R}$. For ease of notation, let each word pair $(x_l, x_r)$ in these two sets correspond to a vector pair $(\mathbf{x}_l, \mathbf{x}_r)$, so that a mini-batch of $k_1$ word pairs is given by $\mathcal{B}_{A} = \lbrack (\mathbf{x}_{l}^{1}, \mathbf{x}_{r}^{1}), \ldots, (\mathbf{x}_{l}^{k_{1}}, \mathbf{x}_{r}^{k_1})\rbrack$ (similarly for $\mathcal{B}_{R}$, which consists of $k_2$ example pairs).

Next, the sets of pseudo-negative examples $T_{A} = \lbrack (\mathbf{t}_{l}^{1}, \mathbf{t}_{r}^{1}), \ldots, (\mathbf{t}_{l}^{k_{1}}, \mathbf{t}_{r}^{k_1})\rbrack$ and $T_{R} = \lbrack (\mathbf{t}_{l}^1, \mathbf{t}_{r}^1), \ldots, (\mathbf{t}_{l}^{k_2}, \mathbf{t}_{r}^{k_2})\rbrack $ are defined as pairs of \emph{negative examples} for each \textsc{Attract} and \textsc{Repel} example pair in mini-batches $\mathcal{B}_{A}$ and $\mathcal{B}_{R}$. These negative examples are chosen from the word vectors present in $\mathcal{B}_A$ or $\mathcal{B}_R$ so that, for each \textsc{Attract} pair $(\mathbf{x}_l, \mathbf{x}_r)$, the negative example pair $(\mathbf{t}_l, \mathbf{t}_r)$ is chosen so that $\mathbf{t}_l$ is the vector closest (in terms of cosine distance) to $\mathbf{x}_l$ and $\mathbf{t}_r$ is closest to $\mathbf{x}_r$. Similarly, for each \textsc{Repel} pair $(\mathbf{x}_l, \mathbf{x}_r)$, the negative example pair $(\mathbf{t}_l, \mathbf{t}_r)$ is chosen from the remaining in-batch vectors so that $\mathbf{t}_l$ is the vector furthest away from $\mathbf{x}_l$ and $\mathbf{t}_r$ is furthest from $\mathbf{x}_r$.

The negative examples are used to: \textbf{a)} force \textsc{Attract} pairs to be closer to each other than to their respective negative examples; and \textbf{b)} to force \textsc{Repel} pairs to be further away from each other than from their negative examples. The first term of the cost function pulls \textsc{Attract} pairs together:
\begin{align} Att(& \mathcal{B}_A, T_{A}) ~=  \notag \\
\sum_{ i = 1}^{k_1} \big[~ &\tau \left( \delta_{att} +  cos(\mathbf{x}_l^{i}, \mathbf{t}_l^{i}) - cos(\mathbf{x}_l^{i},  \mathbf{x}_r^{i}) \right)  \notag \\
+&\tau \left( \delta_{att} +  cos(\mathbf{x}_r^{i}, \mathbf{t}_r^{i}) - cos(\mathbf{x}_l^{i}, \mathbf{x}_r^{i})  \right) \big]
\end{align}
\noindent where $cos$ denotes cosine similarity, $\tau(x)=\max(0,x)$ is the hinge loss function and $\delta_{att}$ is the attract margin which determines how much closer these vectors should be to each other than to their respective negative examples. The second part of the cost function pushes \textsc{Repel} word pairs away from each other:
\begin{align} Rep(& \mathcal{B}_R, {T}_R) ~= \notag \\
\sum_{ i = 1 }^{k_2}  \big[ ~ &\tau\left(\delta_{rep} + cos(\mathbf{x}_l^{i}, \mathbf{x}_r^{i})  - cos(\mathbf{x}_l^{i},\mathbf{t}_l^{i}) \right) \notag \\
+&\tau\left( \delta_{rep} + cos(\mathbf{x}_l^{i}, \mathbf{x}_r^{i})  - cos(\mathbf{x}_r^{i}, \mathbf{t}_r^{i})  \right)  \big] 
\end{align}

\noindent In addition to these two terms, an additional regularisation term is used to \emph{preserve} the abundance of high-quality semantic content present in the distributional vector space, as long as this information does not contradict the injected linguistic constraints. If $V(\mathcal{B})$ is the set of all word vectors present in the given mini-batch, then:
\begin{equation*}
  Reg(\mathcal{B}_A, \mathcal{B}_R) =  \sum\limits_{ \mathbf{x}_i \in V(\mathcal{B}_A \cup \mathcal{B}_R) }  \lambda_{reg} \left\| \widehat{\mathbf{x}_{i}} - \mathbf{x}_i \right\|_{2} 
 \end{equation*}
\noindent where $\lambda_{reg}$ is the L2 regularization constant and $\widehat{\mathbf{x}_{i}}$ denotes the original (distributional) word vector for word $x_i$. The full \textsc{Attract-Repel} cost function is given by the sum of all three terms.

\subsection{\textsc{LEAR}: Encoding Lexical Entailment}
\label{ss:encoding}

In this section, the \textsc{Attract-Repel} framework is extended to model lexical entailment jointly with (symmetric) semantic similarity. To do this, the method uses an additional source of external lexical knowledge: let $L$ be the set of \emph{directed} lexical entailment constraints such as \textit{(corgi, dog)}, \textit{(dog, animal)}, or \textit{(corgi, animal)}, with lower-level concepts on the left and higher-level ones on the right (the source of these constraints will be discussed in Section 3). The optimisation proceeds in the same way as before, considering a mini-batch of \textsc{le} pairs $\mathcal{B}_L$ consisting of $k_3$ word pairs standing in the (directed) lexical entailment relation. 

Unlike symmetric similarity, lexical entailment is an asymmetric relation which encodes a hierarchical ordering between concepts. Inferring the direction of the entailment relation between word vectors requires the use of an asymmetric distance function. We define three different ones, all of which use the word vector's norms to impose an ordering between high- and low-level concepts: 
\begin{align}
 D_{1}(\mathbf{x}, \mathbf{y}) ~~&=~~ ~~~~ |\mathbf{x}| - |\mathbf{y}| \\
 \vspace{2mm}
 D_{2}(\mathbf{x}, \mathbf{y}) ~~&=~~ ~~~~ \frac{|\mathbf{x}| - |\mathbf{y}|}{|\mathbf{x}| + |\mathbf{y}|} \\
 \vspace{2mm}
D_{3}(\mathbf{x}, \mathbf{y}) ~~&=~~ \frac{|\mathbf{x}| - |\mathbf{y}|}{\max(|\mathbf{x}|, |\mathbf{y}|)} 
\end{align}

The lexical entailment term (for the $j$-th asymmetric distance, $j \in 1,2,3$) is  defined as:
\begin{align}
 LE_{j}(\mathcal{B}_L) ~= \sum_{ i = 1 }^{k_3}    D_{j}(\mathbf{x}_{i}, \mathbf{y}_{i})
\label{eq:lej}
\end{align}

The first distance serves as the baseline: it uses the word vectors' norms to order the concepts, that is to decide which of the words is likely to be the higher-level concept. In this case, the magnitude of the difference between the two norms determines the `intensity' of the \textsc{le} relation. This is potentially problematic, as this distance does not impose a limit on the vectors' norms. The second and third metric take a more sophisticated approach, using the ratios of the differences between the two norms and either: \textbf{a)} the sum of the two norms; or \textbf{b)} the larger of the two norms. In doing that, these metrics ensure that the cost function only considers the norms' ratios. This means that the cost function no longer has the incentive to increase word vectors' norms past a certain point, as the magnitudes of norm ratios grow in size much faster than the linear relation defined by the first distance function.

To model the semantic and the \textsc{le} relations jointly, the \textsc{lear} cost function jointly optimises the four terms of the expanded cost function: 
\begin{align*}
C(&\mathcal{B}_A, T_A, \mathcal{B}_R, T_R, \mathcal{B}_L, T_L) ~  = ~  Att(\mathcal{B}_S, T_S) + \ldots ~~~~~ \\ 
 &+~ Rep(\mathcal{B}_A, T_A)  ~+~ Reg(\mathcal{B}_A, \mathcal{B}_R, \mathcal{B}_L) ~+~ \ldots  ~~~~~~~~ \\
&+~ Att(\mathcal{B}_L, T_L) ~+~  LE_{j}(B_{L}) ~~~~~~~~~~~~~~ ~~~~~~~~~~~
\end{align*}
\paragraph{\textsc{LE} Pairs as \textsc{Attract} Constraints} The combined cost function makes use of the batch of lexical constraints $\mathcal{B}_L$ twice: once in the defined asymmetric cost function $LE_{j}$, and once in the symmetric \textsc{Attract} term $Att(\mathcal{B}_L, T_L)$. This means that words standing in the lexical entailment relation are forced to be similar both in terms of cosine distance (via the symmetric \textsc{Attract} term) and in terms of the asymmetric $LE$ distance from Eq.~\eqref{eq:lej}.

\paragraph{Decoding Lexical Entailment}

The defined cost function serves to encode semantic similarity and \textsc{le} relations in the same vector space. Whereas the similarity can be inferred from the standard cosine distance, the \textsc{lear} optimisation embeds lexical entailment as a combination of the symmetric \textsc{Attract} term and the newly defined asymmetric $LE_j$ cost function. Consequently, the metric used to determine whether two words stand in the \textsc{le} relation must combine the two cost terms as well. We define the \textsc{le} \emph{decoding} metric as: 
\begin{align}
I_{LE}(\mathbf{x}, \mathbf{y}) = dcos(\mathbf{x}, \mathbf{y}) + D_{j}(\mathbf{x}, \mathbf{y})
\label{eq:decoding}
\end{align}
where $dcos(\mathbf{x}, \mathbf{y})$ denotes the cosine distance. This decoding function combines the symmetric and the asymmetric cost term, in line with the combination of the two used to perform \textsc{lear} specialisation. In the evaluation, we show that combining the two cost terms has a synergistic effect, with both terms contributing to stronger performance across all \textsc{le} tasks used for evaluation.       

\section{Experimental Setup}
\label{s:experimental}
\paragraph{Starting Distributional Vectors}

To test the robustness of \textsc{lear} specialisation, we experiment with a variety of well-known, publicly available English word vectors: \textbf{1)} Skip-Gram with Negative Sampling (\textsc{SGNS}) \cite{Mikolov:2013nips} trained on the Polyglot Wikipedia \cite{AlRfou:2013conll} by \newcite{Levy:2014acl}; \textbf{2)} \textsc{Glove} Common Crawl \cite{Pennington:2014emnlp}; \textbf{3)} \textsc{context2vec} \cite{Melamud:2016conll}, which replaces CBOW contexts with contexts based on bidirectional LSTMs \cite{Hochreiter:1997}; and \textbf{4)} \textsc{fastText} \cite{Bojanowski:2017tacl}, a SGNS variant which builds word vectors as the sum of their constituent character n-gram vectors.\footnote{All vectors are $300$-dimensional except for the $600$-dimensional \textsc{context2vec } vectors; for further details regarding the architectures and training setup of the used vector collections, we refer the reader to the original papers. We also experimented with dependency-based SGNS vectors \cite{Levy:2014acl}, observing similar patterns in the results.}

\paragraph{Linguistic Constraints} 
We use three groups of linguistic constraints in the \textsc{lear} specialisation model, covering three different relation types which are all beneficial to the specialisation process: directed \textbf{1)} \textit{lexical entailment} (\textsc{le}) \textit{pairs}; \textbf{2)} \textit{synonymy pairs}; and \textbf{3)} \textit{antonymy pairs}. Synonyms are included as symmetric \textsc{attract} pairs (i.e., the $\mathcal{B}_A$ pairs) since they can be seen as defining a trivial symmetric \textsc{is-a} relation \cite{Rei:2014conll,Vulic:2017cl}. For a similar reason, antonyms are clear \textsc{repel} constraints as they anti-correlate with the LE relation.\footnote{In short, the question ``\textit{Is $X$ a type of $X$?}'' (synonymy) is trivially true, while the question ``\textit{Is $\neg X$ a type of $X$?}'' (antonymy) is trivially false.} Synonymy and antonymy constraints are taken from prior work \cite{Zhang:2014emnlp,Ono:2015naacl}: they are extracted from WordNet \cite{Fellbaum:1998wn} and Roget \cite{Kipfer:2009book}. In total, we work with 1,023,082 synonymy pairs (11.7 synonyms per word on average) and 380,873 antonymy pairs (6.5 per word).\footnote{https://github.com/tticoin/AntonymDetection}

As in prior work \cite{Nguyen:2017emnlp,Nickel:2017arxiv}, \textsc{le} constraints are extracted from the WordNet hierarchy, relying on the transitivity of the \textsc{le} relation. This means that we include both direct and indirect \textsc{le} pairs in our set of constraints (e.g., (\textit{pangasius, fish}), (\textit{fish, animal}), and \emph{(pangasius, animal)}). We retained only noun-noun and verb-verb pairs, while the rest were discarded: the final number of \textsc{le} constraints is 1,545,630.\footnote{We also experimented with additional 30,491 \textsc{le} constraints from the Paraphrase Database (PPDB) 2.0 \cite{Pavlick:2015acl}. Adding them to the WordNet-based \textsc{le} pairs makes no significant impact on the final performance. We also used synonymy and antonymy pairs from other sources, such as word pairs from PPDB used previously by \newcite{Wieting:2015tacl}, and BabelNet \cite{Navigli:12} used by \newcite{Mrksic:2017tacl}, reaching the same conclusions.}

\paragraph{Training Setup}
We adopt the original \textsc{Attract-Repel} model setup without any fine-tuning. Hyperparameter values are set to: $\delta_{att}=0.6$, $\delta_{rep}=0.0$, $\lambda_{reg}=10^{-9}$ \cite{Mrksic:2017tacl}. The models are trained for 5 epochs with the AdaGrad algorithm \cite{Duchi:11}, with batch sizes set to $k_1=k_2=k_3=128$ for faster convergence. 

\section{Results and Discussion}
\label{s:results}
We test and analyse \textsc{lear}-specialised vector spaces in two standard word-level \textsc{le} tasks used in prior work: hypernymy directionality and detection (Section~\ref{ss:detectdirect}) and graded \textsc{le} (Section~\ref{ss:gradedle}).

\subsection{LE Directionality and Detection}
\label{ss:detectdirect}
The first evaluation uses three classification-style tasks with increased levels of difficulty. The tasks are evaluated on three datasets used extensively in the \textsc{le} literature \cite{Roller:2014coling,Santus:2014eacl,Weeds:2014coling,Shwartz:2017eacl,Nguyen:2017emnlp}, compiled into an integrated evaluation set by \newcite{Kiela:2015acl}.\footnote{http://www.cl.cam.ac.uk/$\sim$dk427/generality.html}

\begin{figure*}[t]
    \centering
    \begin{subfigure}[t]{0.322\linewidth}
        \centering
        \includegraphics[width=1.0\linewidth]{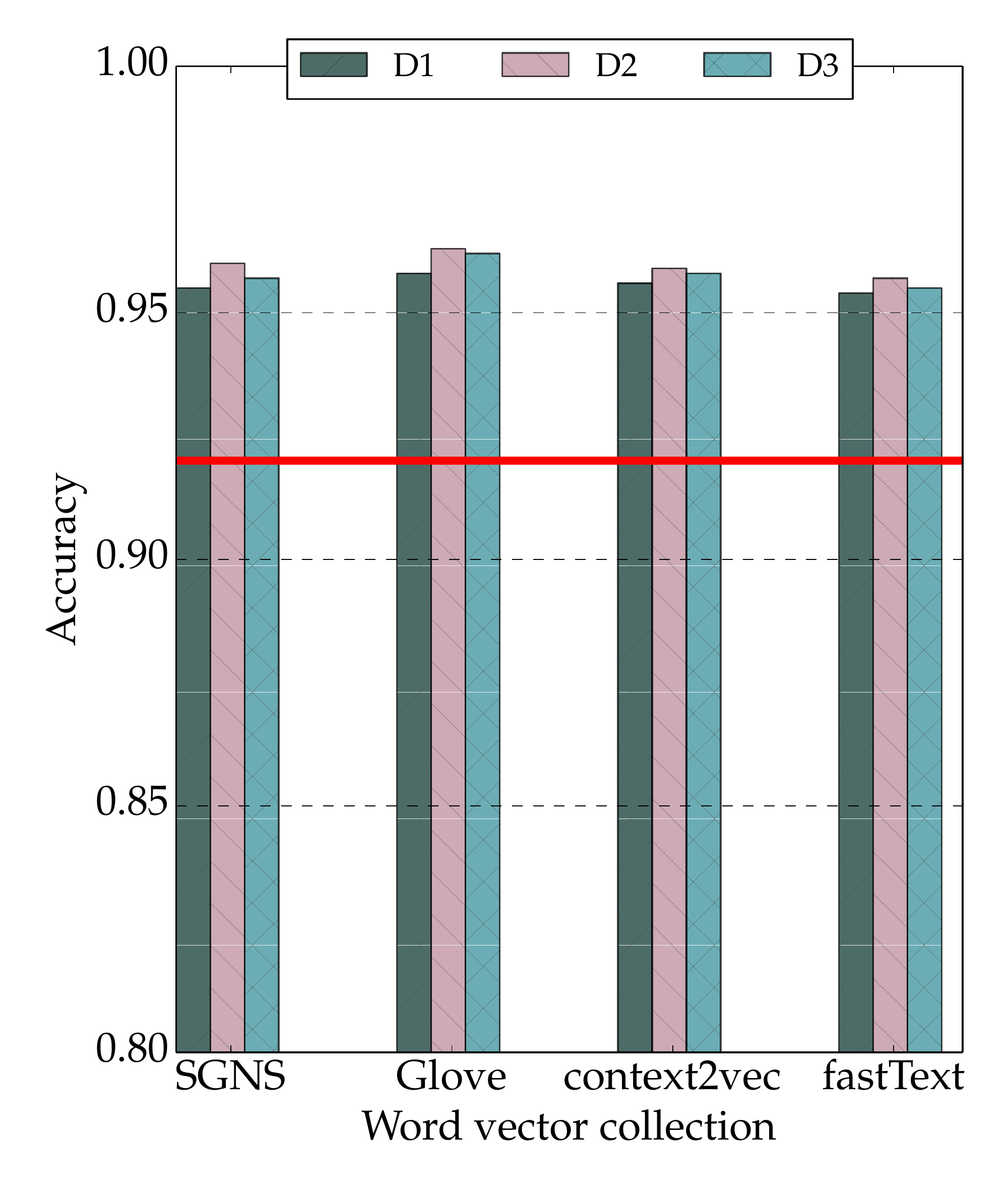}
        \caption{\textbf{BLESS}: Directionality}
        \label{fig:bow}
    \end{subfigure}
    \begin{subfigure}[t]{0.322\textwidth}
        \centering
        \includegraphics[width=1.00\linewidth]{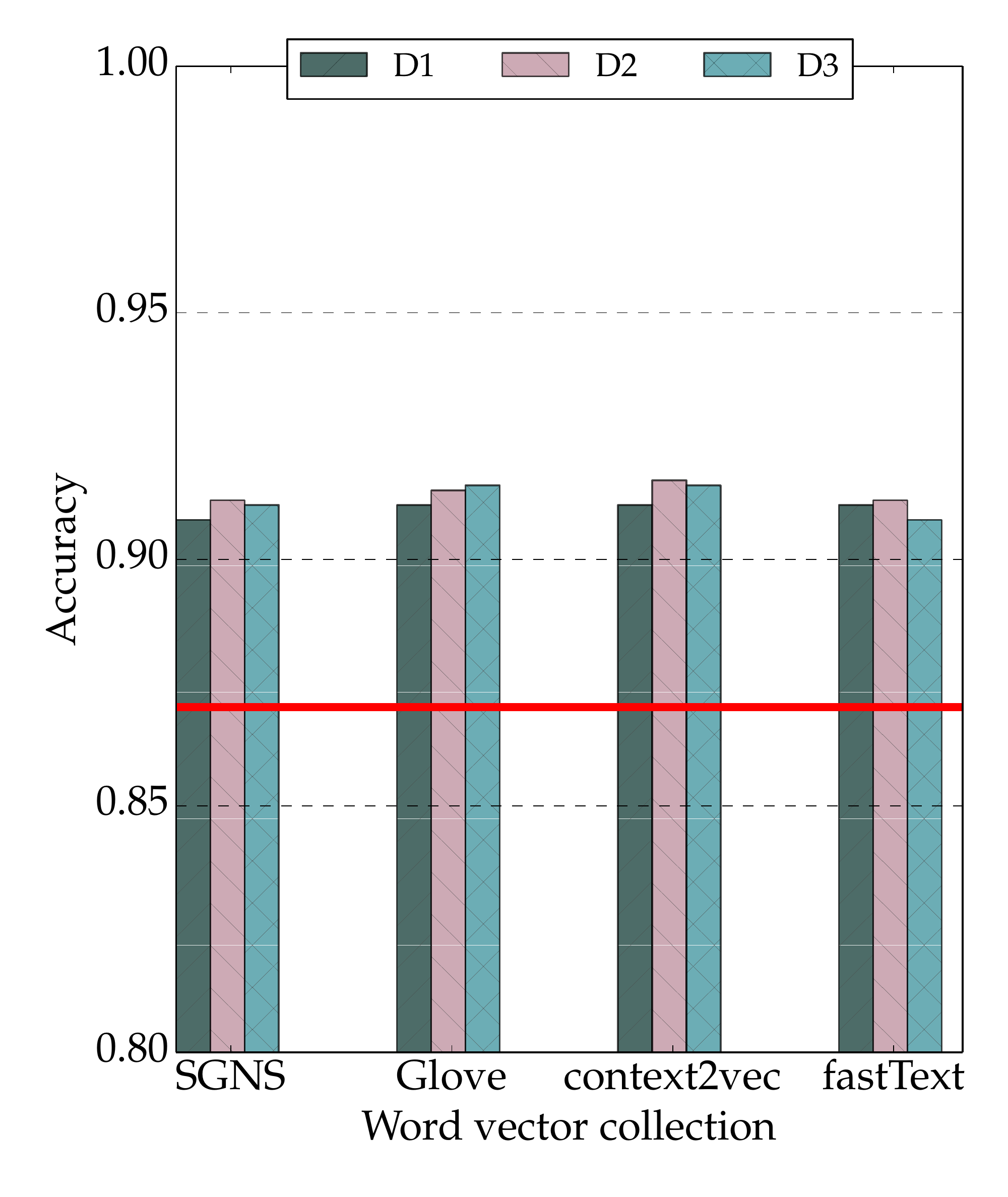}
        \caption{\textbf{WBLESS}: Detection}
        \label{fig:deps}
    \end{subfigure}
    \begin{subfigure}[t]{0.322\textwidth}
        \centering
        \includegraphics[width=1.00\linewidth]{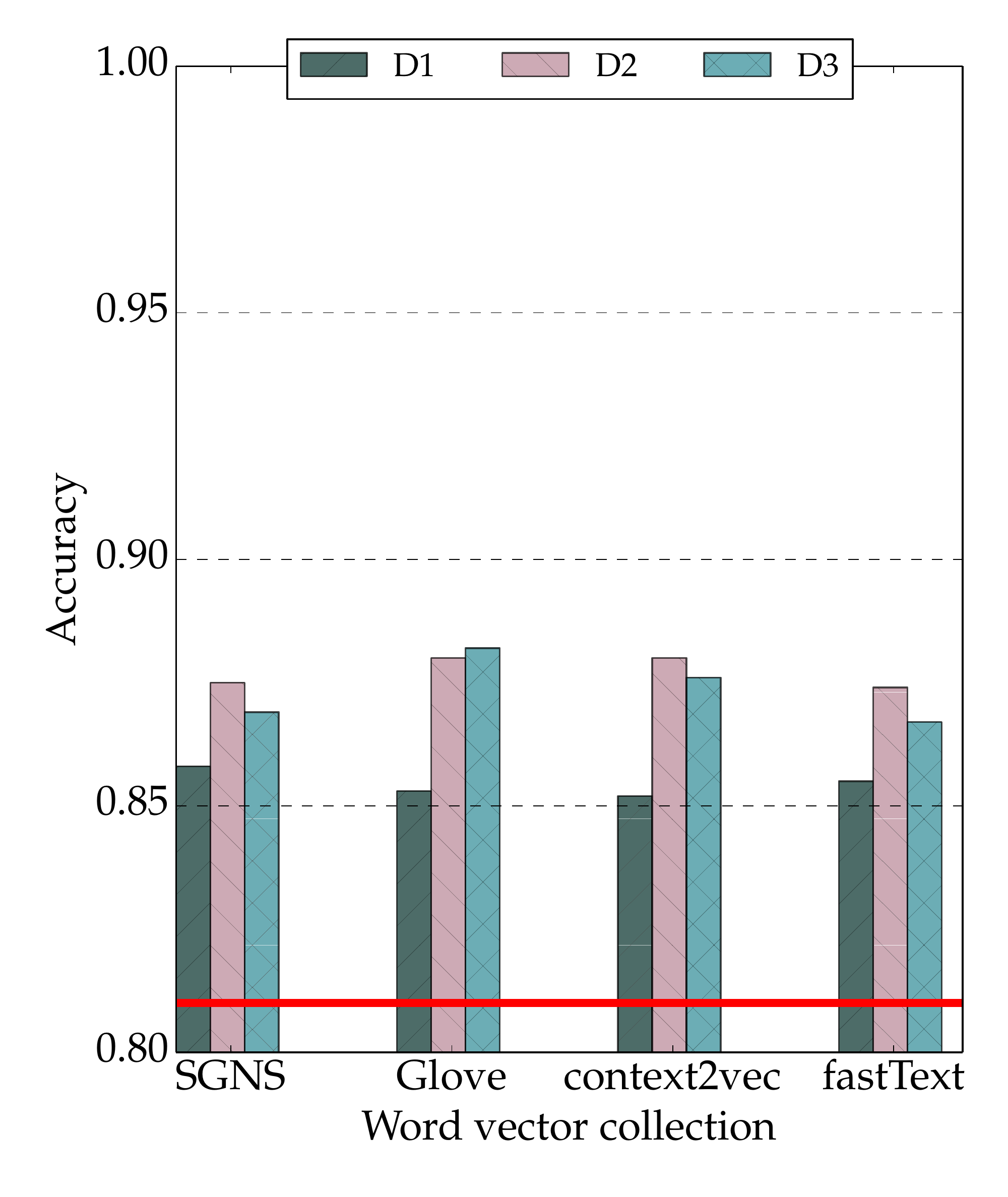}
        \caption{\textbf{BIBLESS}: Detect+direct}
        \label{fig:deps}
    \end{subfigure}
    \vspace{-0.5em}
	\caption{Summary of the results on three different word-level LE subtasks: (a) \textit{directionality}; (b) \textit{detection}; (c) \textit{detection and directionality}. Vertical bars denote the results obtained by different input word vector spaces which are post-processed/specialised by our \textsc{lear} specialisation model using three variants of the asymmetric distance ($D_1$, $D_2$, $D_3$), see Section~\ref{s:methodology}. Thick horizontal red lines refer to the best reported scores on each subtask for these datasets; the baseline scores are taken from \newcite{Nguyen:2017emnlp}.}
\vspace{-0.0mm}
\label{fig:main}
\end{figure*}

The first task, \textsc{le} directionality, is conducted on 1,337 \textsc{le} pairs originating from the \textsc{bless} evaluation set \cite{Baroni:2011bless}. Given a true \textsc{le} pair, the task is to predict the correct hypernym. With \textsc{lear}-specialised vectors this is achieved by simply comparing the vector norms of each concept in a pair: the one with the larger norm is the hypernym (see Figure~\ref{fig:vectors}).

The second task, \textsc{le} detection, involves a binary classification on the \textsc{wbless} dataset \cite{Weeds:2014coling} which comprises 1,668 word pairs standing in a variety of relations (\textsc{le}, meronymy-holonymy, co-hyponymy, reversed \textsc{le}, no relation). The model has to detect a true \textsc{le} pair, that is, to distinguish between the pairs where the statement \textit{$X$ is a (type of) $Y$} is true from all other pairs. With \textsc{lear} vectors, this classification is based on the asymmetric distance score: if the score is above a certain threshold, we classify the pair as ``true \textsc{le}'', otherwise as ``other''. While \newcite{Kiela:2015acl}  manually define the threshold value, we follow the approach of \newcite{Nguyen:2017emnlp} and cross-validate: in each of the 1,000 iterations, 2\% of the pairs are sampled for threshold tuning, and the remaining 98\% are used for testing. The reported numbers are therefore average {accuracy} scores.\footnote{We have conducted more \textsc{le} directionality and detection experiments on other datasets such as EVALution \cite{Santus:2015ws}, the $N_1 \vDash N_2$ dataset of \newcite{Baroni:2012eacl}, and the dataset of \newcite{Lenci:2012sem} with similar performances and findings. We do not report all these results for brevity and clarity of presentation.}

The final task, \textsc{le} detection \textit{and} directionality, concerns a three-way classification on \textsc{bibless}, a relabeled version of \textsc{wbless}. The task is now to distinguish both \textsc{le} pairs ($\rightarrow 1$) and reversed \textsc{le} pairs ($\rightarrow -1$) from other relations ($\rightarrow$ 0), and then additionally select the correct hypernym in each detected \textsc{le} pair. We apply the same test protocol as in the \textsc{le} detection task.


\paragraph{Results and Analysis}
The original paper of \newcite{Kiela:2015acl} reports the following best scores on each task: 0.88 (\textsc{bless}), 0.75 (\textsc{wbless}), 0.57 (\textsc{bibless}). These scores were recently surpassed by \newcite{Nguyen:2017emnlp}, who, instead of post-processing, combine WordNet-based constraints with an SGNS-style objective into a joint model. They report the best scores to date: 0.92 (\textsc{bless}), 0.87 (\textsc{wbless}), and 0.81 (\textsc{bibless}). 

The performance of the four \textsc{lear}-specialised word vector collections is shown in Figure~\ref{fig:main} (together with the strongest baseline scores for each of the three tasks). The comparative analysis confirms the increased complexity of subsequent tasks. \textsc{lear} specialisation of \emph{each} of the starting vector spaces consistently outperformed \emph{all} baseline scores across \emph{all} three tasks. The extent of the improvements is correlated with task difficulty: it is lowest for the easiest directionality task ($0.92 \rightarrow 0.96$), and highest for the most difficult detection plus directionality task ($0.81 \rightarrow 0.88$).

The results show that the two \textsc{lear} variants which do not rely on absolute norm values and perform a normalisation step in the asymmetric distance (D2 and D3) have an edge over the D1 variant which operates with unbounded norms. The difference in performance between D2/D3 and D1 is even more pronounced in the graded LE task (see Section~\ref{ss:gradedle}). This shows that the use of unbounded vector norms diminishes the importance of the symmetric cosine distance in the combined asymmetric distance. Conversely, the synergistic combination used in D2/D3 does not suffer from this issue. 

The high scores achieved with each of the four word vector collections show that \textsc{lear} is not dependent on any particular word representation architecture. Moreover, the extent of the performance improvements in each task suggests that \textsc{lear} is able to reconstruct the concept hierarchy coded in the input linguistic constraints. 

Moreover, we have conducted a small experiment to verify that the \textsc{lear} method can generalise beyond what is directly coded in pairwise external constraints. A simple WordNet lookup baseline yields accuracy scores of 0.82 and 0.80 on the directionality and detection tasks, respectively. This baseline is outperformed by \textsc{lear}: its scores are 0.96 and 0.92 on the two tasks when relying on the same set of WordNet constraints.

\begin{table}[t]
{\scriptsize
        \def\arraystretch{0.99}
        \begin{tabularx}{\linewidth}{lr lr lr}
   		\toprule
        {} & {\textbf{Norm}} &  {} & {\textbf{Norm}}  & {} & {\textbf{Norm}}\\
        \cmidrule(lr){2-2} \cmidrule(lr){3-4} \cmidrule(lr){5-6}
        {terrier} & {0.87} & {laptop} & {0.60} & {cabriolet} & {0.74} \\
        {dog} & {2.64} & {computer} & {2.96} & {car} & {3.59} \\
        {mammal} & {8.57} & {machine} & {6.15} & {vehicle} & {7.78} \\
        {vertebrate} & {10.96}  & {device} & {12.09} & {transport} & {8.01} \\
        {animal} & {11.91} & {artifact} & {17.71} & {instrumentality} & {14.56} \\
        {organism} & {20.08} & {object} & {23.55} & {--} & {--} \\
        \bottomrule
        \end{tabularx}
        }%
        \vspace{-0.5em}
        \caption{L2 norms for selected concepts from the WordNet hierarchy. Input: \textsc{fastText}; \textsc{lear}: D2.}
        \label{tab:norms}
\end{table}

\paragraph{Importance of Vector Norms} To verify that the knowledge concerning the position in the semantic hierarchy actually arises from vector norms, we also manually inspect the norms after \textsc{lear} specialisation. A few examples are provided in Table~\ref{tab:norms}. They indicate a desirable pattern in the norm values which imposes a hierarchical ordering on the concepts. Note that the original distributional SGNS model \cite{Mikolov:2013nips} does not normalise vectors to unit length after training. However, these norms are not at all correlated with the desired hierarchical ordering, and are therefore useless for \textsc{le}-related applications: the non-specialised distributional SGNS model scores 0.44, 0.48, and 0.34 on the three tasks, respectively.
\begin{figure*}[t]
    \centering
    \subcaptionbox{HyperLex: \textbf{All}}{
        \includegraphics[scale=0.27]{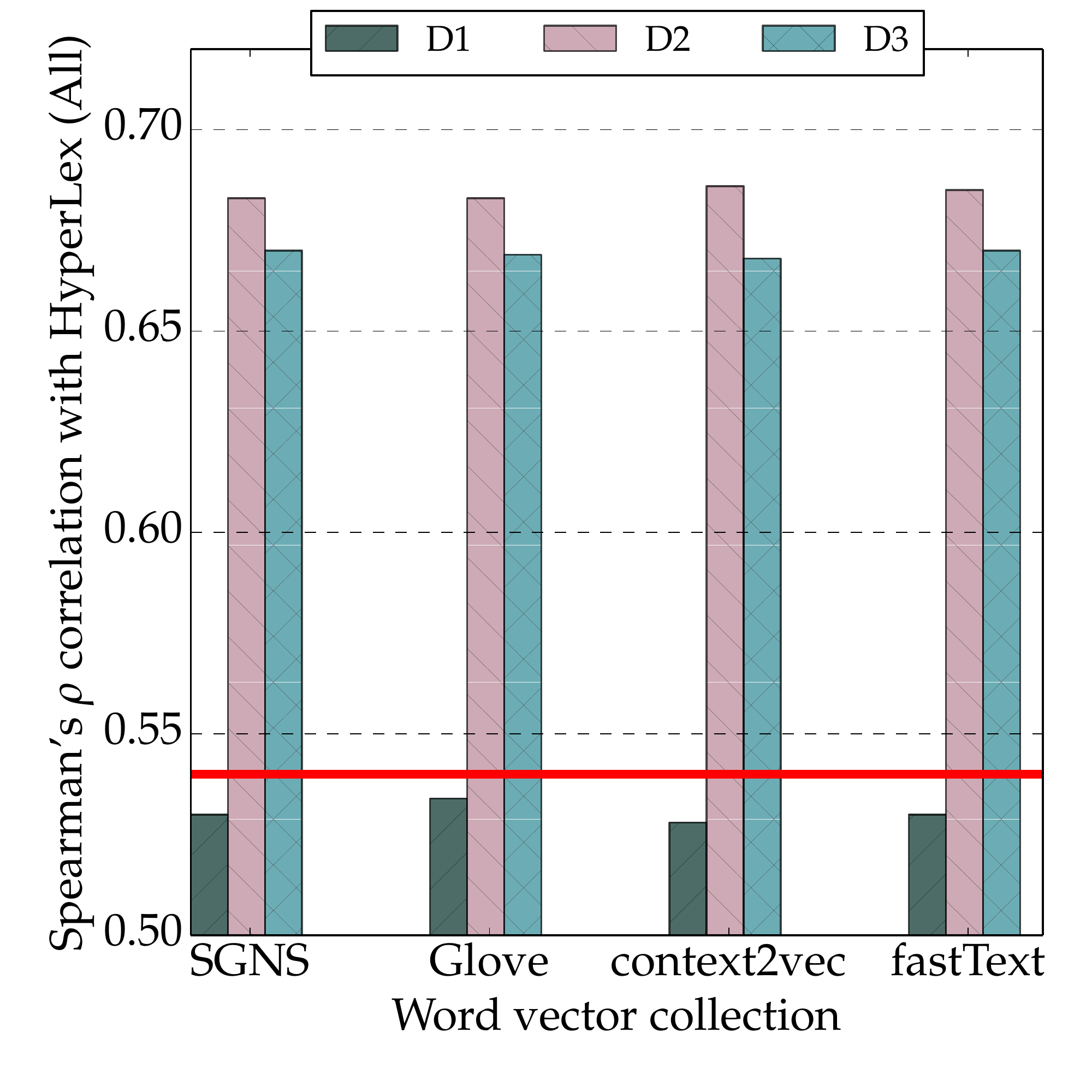}
    }
    \subcaptionbox{HyperLex: \textbf{Nouns}}{
        \includegraphics[scale=0.27]{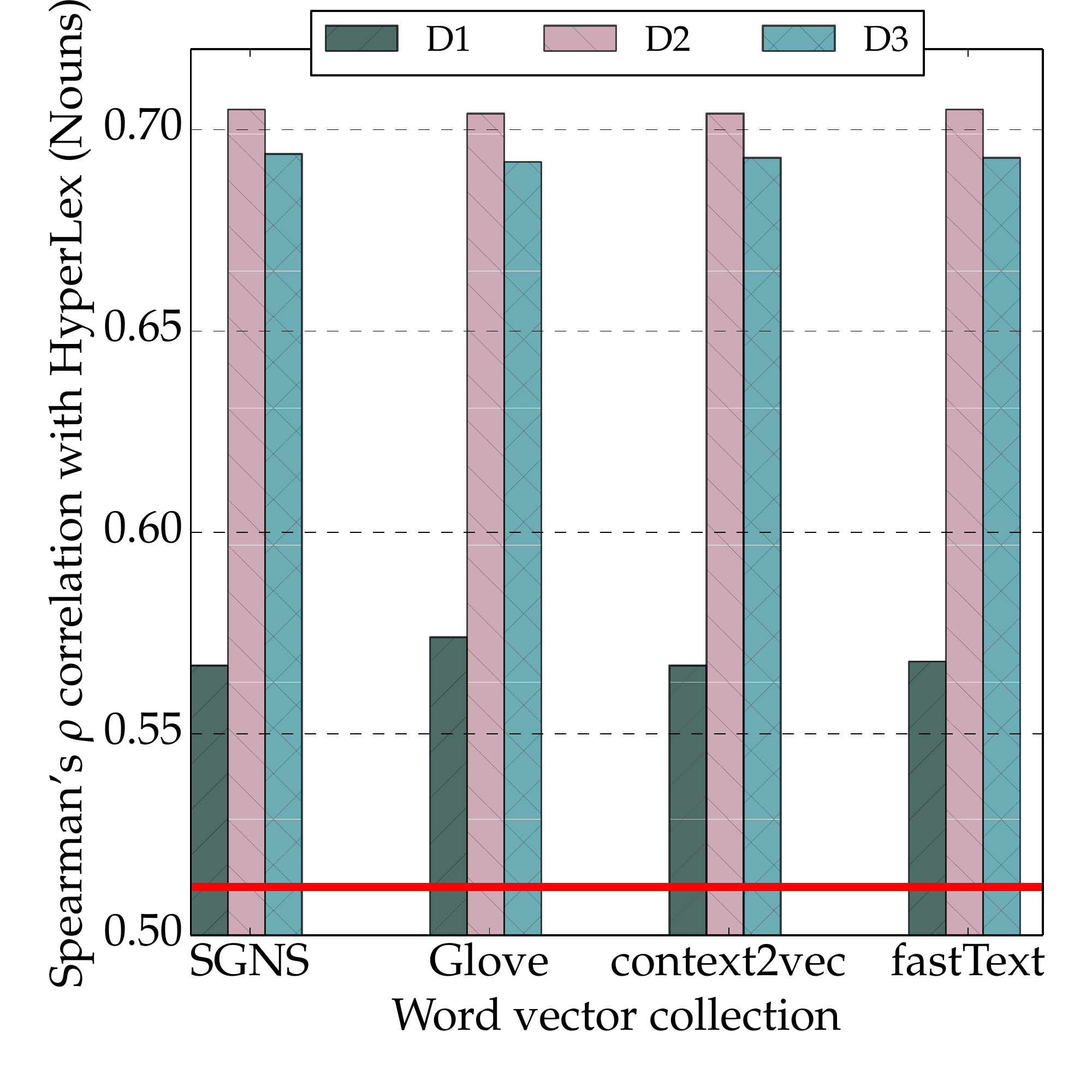}
    }
    \subcaptionbox[]{\textbf{Summary}}{
        {\footnotesize
        \def\arraystretch{0.98}
        \begin{tabular}{l c}
        	\toprule
            {} & All \\
            \cmidrule(lr){2-2}
            {\textsc{freq-ratio}} & 0.279 \\
            {\textsc{sgns (cos)}} & 0.205 \\
            {\textsc{slqs-sim}} & 0.228 \\
            {\textsc{visual}} & 0.209 \\
            {\textsc{wn-best}} & 0.234 \\
         	{\textsc{word2gauss}} & 0.206 \\
            {\textsc{sim-spec}} & 0.320 \\
            \midrule
            {\textsc{order-emb}} & 0.191 \\
            {\textsc{Poincar\'{e}} (nouns)} & {0.512} \\
            {\textsc{HyperVec}} & {0.540} \\
            \midrule
            {{Best} \textsc{lear}} & {\bf 0.686} \\
            \bottomrule
        \end{tabular}
        \vspace{5.5mm}
        }%
    }
    \caption{Results on the graded \textsc{LE} task defined by HyperLex. Following \newcite{Nickel:2017arxiv}, we use Spearman's rank correlation scores on: \textbf{a)} the entire dataset (2,616 noun and verb pairs); and \textbf{b)} its noun subset (2,163 pairs). The summary table shows the performance of other well-known architectures on the full HyperLex dataset, compared to the best results achieved using \textsc{lear} specialisation.}
    \label{fig:gradedle}
\end{figure*}

\subsection{Graded Lexical Entailment}
\label{ss:gradedle}
Asymmetric distances in the \textsc{lear}-specialised space quantify the degree of lexical entailment between any two concepts. This means that they can be used to make fine-grained assertions regarding the hierarchical relationships between concepts. We test this property on HyperLex \cite{Vulic:2017cl}, a gold standard dataset for evaluating how well word representation models capture graded \textsc{le}, grounded in the notions of \textit{concept (proto)typicality} \cite{Rosch:1973:natural,Medin:1984jep} and \textit{category vagueness} \cite{Kamp:1995cog,Hampton:2007cogsci} from cognitive science. HyperLex contains 2,616 word pairs (2,163 noun pairs and 453 verb pairs) scored by human raters in the $[0,6]$ interval following the question \textit{``To what degree is X a (type of) Y?''}\footnote{From another perspective, one might say that graded \textsc{le} provides finer-grained human judgements on a continuous scale rather than simplifying the judgements into binary discrete decisions. For instance, the HyperLex score for the pair \textit{(girl, person)} is 5.91/6, the score for \textit{(guest, person)} is 4.33, while the score for the reversed pair \textit{(person, guest)} is 1.73.}


As shown by the high inter-annotator agreement on HyperLex (0.85), humans are able to consistently reason about graded \textsc{le}.\footnote{For further details concerning HyperLex, we refer the reader to the resource paper \cite{Vulic:2017cl}. The dataset is available at: \url{http://people.ds.cam.ac.uk/iv250/hyperlex.html}} However, current state-of-the-art representation architectures are far from this ceiling. For instance, \newcite{Vulic:2017cl} evaluate a plethora of architectures and report a high-score of only 0.320 (see the summary table in Figure~\ref{fig:gradedle}). Two recent representation models \cite{Nickel:2017arxiv,Nguyen:2017emnlp} focused on the \textsc{LE} relation in particular (and employing the same set of WordNet-based constraints as \textsc{lear}) report the highest score of 0.540 (on the entire dataset) and 0.512 (on the noun subset).


\paragraph{Results and Analysis}
We scored all HyperLex pairs using the combined asymmetric distance described by Equation~\eqref{eq:decoding}, and then computed Spearman’s rank correlation with the ground-truth ranking. Our results, together with the strongest baseline scores, are summarised in Figure~\ref{fig:gradedle}. 


The summary table in Figure~\ref{fig:gradedle}(c) shows the HyperLex performance of several prominent \textsc{le} models. We provide only a quick outline of these models here; further details can be found in the original papers. \textsc{freq-ratio} exploits the fact that more general concepts tend to occur more frequently in textual corpora. \textsc{sgns (cos)} uses non-specialised SGNS vectors and quantifies the \textsc{LE} strength using the symmetric cosine distance between vectors. A comparison of these models to the best-performing \textsc{lear} vectors shows the extent of the improvements achieved using the specialisation approach. 

\textsc{lear}-specialised vectors also outperform \textsc{slqs-sim} \cite{Santus:2014eacl} and \textsc{visual} \cite{Kiela:2015acl}, two \textsc{le} detection models similar in spirit to \textsc{lear}. These models combine symmetric semantic similarity (through cosine distance) with an asymmetric measure of lexical generality obtained either from text (\textsc{slqs-sim}) or visual data (\textsc{visual}). The results on HyperLex indicate that the two generality-based measures are too coarse-grained for graded \textsc{le} judgements. These models were originally constructed to tackle \textsc{le} directionality and detection tasks (see Section~\ref{ss:detectdirect}), but their performance is surpassed by \textsc{lear} on those tasks as well. The \textsc{visual} model outperforms \textsc{slqs-sim}. However, its numbers on \textsc{bless} (0.88), \textsc{wbless} (0.75), and \textsc{bibless} (0.57) are far from the top-performing \textsc{lear} vectors (0.96, 0.92, 0.88).\footnote{We note that \textsc{slqs} and \textsc{visual} do not leverage any external knowledge from WordNet, but the \textsc{visual} model leverages external visual information about concepts.}

\textsc{wn-best} denotes the best result with asymmetric similarity measures which use the WordNet structure as their starting point \cite{Wu:1994acl,Pedersen:2004aaai}. This model can be observed as a model that directly looks up the full WordNet structure to reason about graded lexical entailment. The reported results from Figure~\ref{fig:gradedle}(c) suggest it is more effective to quantify the LE relation strength by using WordNet as the source of constraints for specialisation models such as \textsc{HyperVec} or \textsc{lear}.

\textsc{word2gauss} \cite{Vilnis:2015iclr} represents words as multivariate $K$-dimensional Gaussians rather than points in the embedding space: it is therefore naturally asymmetric and was used in \textsc{le} tasks before, but its performance on HyperLex indicates that it cannot effectively capture the subtleties required to model graded \textsc{le}. However, note that the comparison is not strictly fair as \textsc{word2gauss} does not leverage any external knowledge. An interesting line for future research is to embed external knowledge within this representation framework.

Most importantly, \textsc{lear} outperforms three recent (and conceptually different) architectures: \textsc{order-emb} \cite{Vendrov:2016iclr}, \textsc{Poincar\'{e}} \cite{Nickel:2017arxiv}, and \textsc{HyperVec} \cite{Nguyen:2017emnlp}. Like \textsc{lear}, all of these models complement distributional knowledge with external linguistic constraints extracted from WordNet. Each model uses a different strategy to exploit the hierarchical relationships encoded in these constraints (their approaches are discussed in Section~\ref{s:rw}).\footnote{As discussed previously by \newcite{Vulic:2017cl}, the off-the-shelf \textsc{order-emb} vectors were trained for the binary ungraded LE detection task: this limits their expressiveness in the graded LE task.} However, \textsc{lear}, as the first \textsc{le}-oriented post-processor, is able to utilise the constraints more effectively than its competitors. Another advantage of \textsc{lear} is its applicability to any input vector space.

Figures~\ref{fig:gradedle}(a) and \ref{fig:gradedle}(b) indicate that the two \textsc{lear} variants which rely on norm ratios (D2 and D3), rather than on absolute (unbounded) norm differences (D1), achieve stronger performance on HyperLex. The highest correlation scores are again achieved by D2 with all input vector spaces.

\subsection{Further Discussion}
\label{ss:further}
\paragraph{Why Symmetric + Asymmetric?}
In another experiment, we analyse the contributions of both \textsc{le}-related terms in the \textsc{lear} combined objective function (see Section~\ref{ss:encoding}). We compare three variants of \textsc{lear}: \textbf{1)} a symmetric variant which does not arrange vector norms using the $LE_j(\mathcal{B}_L)$ term (\textsc{sym-only}); \textbf{2)} a variant which arranges norms, but does not use \textsc{le} constraints as additional symmetric \textsc{attract} constraints (\textsc{asym-only}); and \textbf{3)} the full \textsc{lear} model, which uses both cost terms (\textsc{full}). The results with one input space (similar results are achieved with others) are shown in Table~\ref{tab:ablation}. This table shows that, while the stand-alone \textsc{asym-only} term seems more beneficial than the \textsc{sym-only} one, using the two terms jointly yields the strongest performance across all \textsc{LE} tasks.
 \begin{table}[!t]
 {\footnotesize
 \begin{tabularx}{\linewidth}{l XXXX}
 \toprule
 {} & {\textsc{wbless}} & {\textsc{bibless}} & {\textsc{hl-a}} &  {\textsc{hl-n}} \\
 \cmidrule(lr){2-5}
 {\textbf{\textsc{lear} variant}} & {} & {} & {} & {} \\
 {\textsc{sym-only}} & {0.687} & {0.679} & {0.469} & {0.429} \\
 {\textsc{asym-only}} & {0.867} & {0.824} & {0.529} & {0.565} \\
 {\textsc{full}} & {\bf 0.912} & {\bf 0.875} & {\bf 0.686} & {\bf 0.705} \\
 \bottomrule
 \end{tabularx}
 }%
 \caption{Analysing the importance of the synergy in the \textsc{full} \textsc{lear} model on the final performance on \textsc{wbless}, \textsc{bless}, HyperLex-All (\textsc{hl-a}) and HyperLex-Nouns (\textsc{hl-n}). Input: \textsc{fastText}. D2.}
 \vspace{-0.0em}
 \label{tab:ablation}
 \end{table}

\paragraph{\textsc{LE} and Semantic Similarity}
We also test whether the asymmetric $LE$ term harms the (norm-independent) cosine distances used to represent semantic similarity. The \textsc{lear} model is compared to the original \textsc{attract-repel} model making use of the same set of linguistic constraints. Two true semantic similarity datasets are used for evaluation: SimLex-999 \cite{Hill:2015cl} and SimVerb-3500 \cite{Gerz:2016emnlp}. There is no significant difference in performance between the two models, both of which yield similar results on SimLex (Spearman's rank correlation of $\approx$ 0.71) and SimVerb ($\approx$ 0.70). This proves that cosine distances remain preserved during the optimization of the asymmetric objective performed by the joint \textsc{lear} model.

\section{Related Work}
\label{s:rw}
\paragraph{Vector Space Specialisation} A standard approach to incorporating external information into vector spaces is to pull the representations of similar words closer together. Some models integrate such constraints into the training procedure: they modify the prior or the regularisation \cite{Yu:2014,Xu:2014,Bian:14,Kiela:2015emnlp}, or use a variant of the SGNS-style objective \cite{Liu:EtAl:15,Osborne:16,Nguyen:2017emnlp}. Another class of models, popularly termed \textit{retrofitting}, fine-tune distributional vector spaces by injecting lexical knowledge from semantic databases such as WordNet or the Paraphrase Database   \cite{Faruqui:2015naacl,Jauhar:2015,Wieting:2015tacl,Nguyen:2016acl,Mrksic:2016naacl,Mrksic:2017tacl}. 

\textsc{lear} falls into the latter category. However, while previous post-processing methods have focused almost exclusively on specialising vector spaces to emphasise semantic similarity (i.e., to distinguish between similarity and relatedness by explicitly pulling synonyms closer and pushing antonyms further apart), this paper proposed a principled methodology for specialising vector spaces for asymmetric hierarchical relations (of which \textit{lexical entailment} is an instance). Its starting point is the state-of-the-art similarity specialisation framework of \newcite{Mrksic:2017tacl}, which we extend to support the inclusion of hierarchical asymmetric relationships between words.

\paragraph{Word Vectors and Lexical Entailment}
Since the hierarchical \textsc{le} relation is one of the fundamental building blocks of semantic taxonomies and hierarchical concept categorisations \cite{Beckwith:1991wn,Fellbaum:1998wn}, a significant amount of research in semantics has been invested into its automatic detection and classification. Early work relied on asymmetric directional measures \cite[i.a.]{Weeds:2004coling,Clarke:2009gems,Kotlerman:2010nle,Lenci:2012sem} which were based on the distributional inclusion hypothesis \cite{Geffet:2005acl} or the distributional informativeness or generality hypothesis \cite{Herbelot:2013acl,Santus:2014eacl}. However, these approaches have recently been superseded by methods based on word embeddings. These methods build dense real-valued vectors for capturing the \textsc{LE} relation, either directly in the \textsc{LE}-focused space \cite{Vilnis:2015iclr,Vendrov:2016iclr,Henderson:2016acl,Nickel:2017arxiv,Nguyen:2017emnlp} or by using the vectors as features for supervised \textsc{LE} detection models \cite{Tuan:2016emnlp,Shwartz:2016acl,Nguyen:2017emnlp,Glavas:2017emnlp}. 

Several \textsc{le} models embed useful hierarchical relations from external resources such as WordNet into \textsc{le}-focused vector spaces, with solutions coming in different flavours. The model of \newcite{Yu:2015ijcai} is a dynamic distance-margin model optimised for the \textsc{le} detection task using hierarchical WordNet constraints. This model was extended by \newcite{Tuan:2016emnlp} to make use of contextual sentential information. A major drawback of both models is their inability to make directionality judgements. Further, their performance has  recently been surpassed by the \textsc{HyperVec} model of \newcite{Nguyen:2017emnlp}. This model combines WordNet constraints with the SGNS distributional objective into a {joint} model. As such, the model is tied to the SGNS objective and any change of the distributional modelling paradigm implies a change of the entire \textsc{HyperVec} model. This makes their model less versatile than the proposed \textsc{lear} framework. Moreover, the results achieved using \textsc{lear} specialisation achieve substantially better performance across all \textsc{le} tasks used for evaluation.

Another model similar in spirit to \textsc{lear} is the \textsc{order-emb} model of \newcite{Vendrov:2016iclr}, which encodes hierarchical structure by imposing a partial order in the embedding space: higher-level concepts get assigned higher per-coordinate values in a $d$-dimensional vector space. The model minimises the violation of the per-coordinate orderings during training by relying on hierarchical WordNet constraints between word pairs. Finally, the \textsc{Poincar\'{e}} model of \newcite{Nickel:2017arxiv} makes use of hyperbolic spaces to learn general-purpose \textsc{LE} embeddings based on $n$-dimensional Poincar\'{e} balls which encode both hierarchy and semantic similarity, again using the WordNet constraints. A similar model in hyperbolic spaces was proposed by \newcite{Chamberlain:2017arxiv}. In this paper, we demonstrate that \textsc{le}-specialised word embeddings with stronger performance can be induced using a simpler model operating in more intuitively interpretable Euclidean vector spaces.

\section{Conclusion and Future Work}
\label{s:conclusion}
This paper proposed \textsc{lear}, a vector space specialisation procedure which simultaneously injects symmetric and asymmetric constraints into existing vector spaces, performing joint specialisation for two properties: \emph{lexical entailment} and \emph{semantic similarity}. Since the former is not symmetric, \textsc{lear} uses an asymmetric cost function which encodes the hierarchy between concepts by manipulating the norms of word vectors, assigning higher norms to higher-level concepts. Specialising the vector space for both relations has a synergistic effect: \textsc{lear}-specialised vectors attain state-of-the-art performance in judging semantic similarity and set new high scores across four different lexical entailment tasks. The code for the \textsc{lear} model is available from: \url{github.com/nmrksic/lear}.

In future work, we plan to apply a similar methodology to other asymmetric relations (e.g.,~\emph{meronymy}), as well as to investigate fine-grained models which can account for differing path lengths from the WordNet hierarchy. We will also extend the model to reason over words unseen in input lexical resources, similar to the recent post-specialisation model oriented towards specialisation of unseen words for similarity \cite{Vulic:2018naaclpost}.  We also plan to test the usefulness of LE-specialised vectors in downstream natural language understanding tasks. Porting the model to other languages and enabling cross-lingual applications such as cross-lingual lexical entailment \cite{Upadhyay:2018naacl} is another future research direction.

\section*{Acknowledgments}
This work is supported by the ERC Consolidator Grant LEXICAL: Lexical Acquisition Across Languages (no 648909). NM performed his work while he was still at the University of Cambridge. We are grateful to the TakeLab research group at the University of Zagreb, especially to Mladen Karan, for offering support to computationally intensive experiments in our hour of need.

\bibliography{acl2017_refs}
\bibliographystyle{emnlp_natbib}

\end{document}